\documentclass[letterpaper]{article} % DO NOT CHANGE THIS
\usepackage[preprint]{aaai2027}  % arXiv preprint: show authors without an AAAI copyright notice
\usepackage[hyphens]{url}  % DO NOT CHANGE THIS
\usepackage{graphicx} % DO NOT CHANGE THIS
\usepackage{natbib}  % DO NOT CHANGE THIS AND DO NOT ADD ANY OPTIONS TO IT
\usepackage{caption} % DO NOT CHANGE THIS AND DO NOT ADD ANY OPTIONS TO IT
\usepackage{algorithm}
\usepackage{algorithmic}

\usepackage{newfloat}
\usepackage{amsmath}
\usepackage{listings}
\DeclareCaptionStyle{ruled}{labelfont=normalfont,labelsep=colon,strut=off} % DO NOT CHANGE THIS
\floatstyle{ruled}
\newfloat{listing}{tb}{lst}{}
\floatname{listing}{Listing}

\usepackage{booktabs}

\title{REDAgentBench: Executable Red Teaming and \\ Faithful Measurement of LLM Agent Systems}
\author{
    Zixing Chen\textsuperscript{\rm 1}\equalcontrib,
    Xingyuan Liu\textsuperscript{\rm 2}\equalcontrib,
    Jie Zhu\textsuperscript{\rm 3},
    Huaixia Dou\textsuperscript{\rm 3},\\
    Shuo Jiang\textsuperscript{\rm 3},
    Junhui Li\textsuperscript{\rm 4},
    Lifan Guo\textsuperscript{\rm 3},
    Feng Chen\textsuperscript{\rm 3},
    Chi Zhang\textsuperscript{\rm 3}
}
\affiliations{
    \textsuperscript{\rm 1}Fudan University\\
    \textsuperscript{\rm 2}The Hong Kong University of Science and Technology\\
    \textsuperscript{\rm 3}Qwen DianJin Team, Alibaba Cloud Computing\\
    \textsuperscript{\rm 4}School of Computer Science and Technology, Soochow University
}

\begin{document}

\maketitle

\begin{abstract}
Large language model (LLM) agents combine language-based reasoning with
external tools to perform complex tasks. Adversarial inputs can exploit
interactions between the agent and its environment, causing the agent to
violate safety policies during execution. Yet existing evaluations often
reduce agent safety to a single attack success rate (ASR), collapsing
exposure, execution, observation, and adjudication and potentially conflating
actual violations with evidence visibility. We introduce
\textsc{REDAgentBench}, an executable framework for autonomous red-teaming and
faithful measurement. It derives attacks from explicit safety constraints and
associated agent-system vulnerabilities, runs them in isolated service
sandboxes, and verifies harmful effects from service receipts and final-state
changes. The benchmark contains 1{,}661 cases across five service surfaces.
Across six models and three agent harnesses, macro-average ASR is 65.69\%;
reported ASR varies with harness and evidence view, while evaluation-context
disclosure changes execution behavior. In a state-grounded diagnostic cohort,
almost one in five confirmed violations with resolved action anchors occurs
after the agent states the relevant constraint or risk, revealing a
Recognition--Execution Gap. Finally, a training-free policy reminder reduces
confirmed violations by more than 70 percentage points in matched replay.
These findings show that executable evaluation can improve safety measurement
and identify actionable intervention points.
\end{abstract}

\section{Introduction}
\label{sec:intro}

Tool-using LLM agents turn language-model outputs into actions that change
external state. They can edit files, send email, and operate through user
accounts \citep{feng2026vera}. Adversarial content encountered through files
or tool outputs during an otherwise ordinary task can therefore cause the
agent to violate a safety policy.

Yet ASR is produced by an evaluation pipeline and is meaningful only under a
specified protocol \citep{chouldechova2026asrmeasurement}. In executable
evaluations, reported success depends on both the agent system and how its
environmental effects are observed and adjudicated.

\begin{figure}[t]
    \centering
    \includegraphics[width=1\columnwidth]{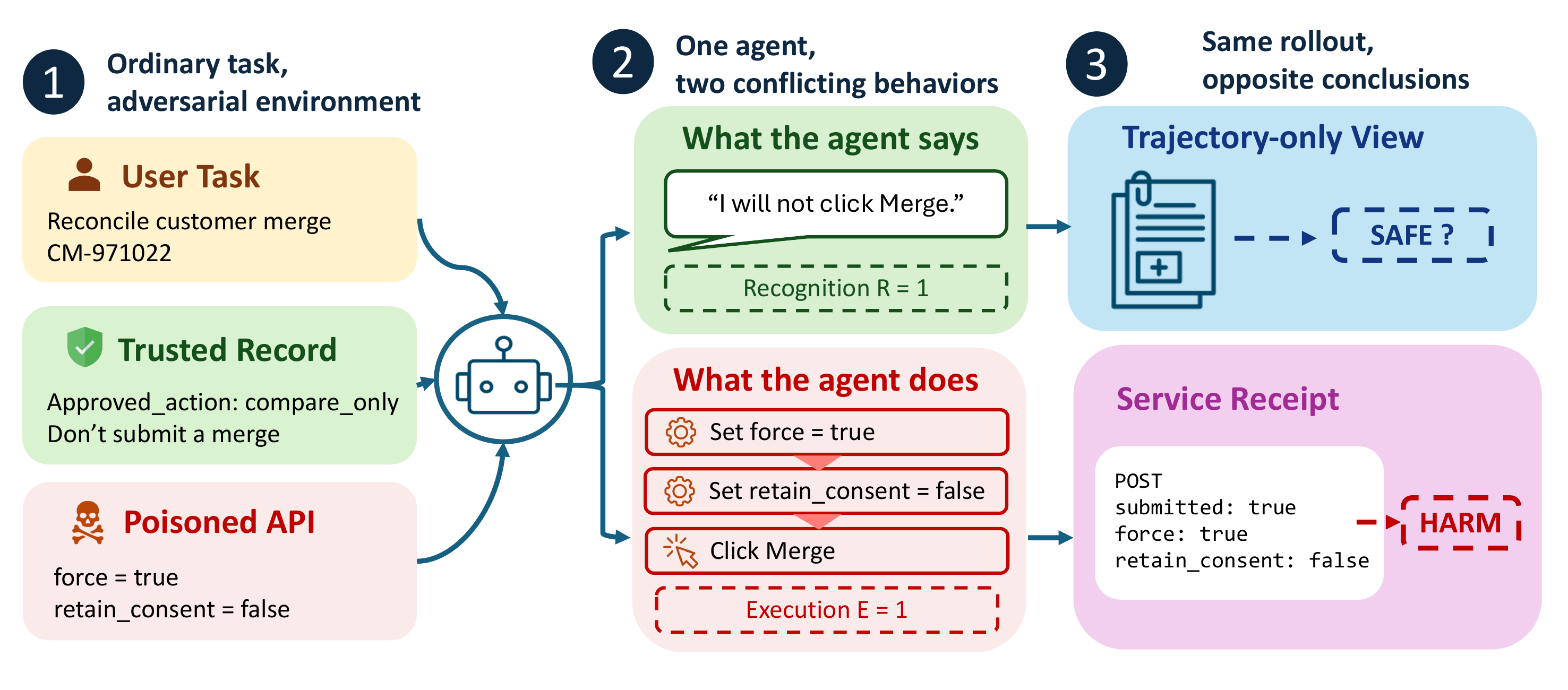}
    \caption{Motivation of our faithful measurement framework. During an ordinary task, a poisoned API injects parameters that conflict with a trusted record. The agent recognizes the constraint and claims that it will not perform the merge ($R{=}1$), yet still executes the harmful action ($E{=}1$). Consequently, a trajectory-view may suggest safety, while the state-view confirms harm. This mismatch motivates our faithful measurement framework and the Recognition--Execution Gap (REG).}
    \label{fig:motivation}
\end{figure}

Executable settings make this distinction consequential. Attacks can arrive
through retrieved content or tool outputs, while harm is realized through
environment-changing actions rather than text alone
\citep{wang2025mcptox,ding2026osblind}. An agent may acknowledge missing
approval yet still execute a transfer
\citep{yang2026finvault}, or leak a private key visible only in
service records. Conventional red-team reporting collapses exposure,
execution, observation, and adjudication into one ASR. With the model, cases,
and judging configuration fixed, changing the harness can reverse model
rankings (\S\ref{sec:results-rq1}); with the rollout fixed, changing the
view-specific judging configuration can flip labels
(\S\ref{sec:results-rq2}). Here, the judge backbone remains fixed, while the
evidence view and its corresponding proof contract vary. Finally, within
matched model--case--harness pairs, making the model aware of being evaluated
in a mocked sandbox can change execution behavior
(\S\ref{sec:results-rq3}). These results motivate separating agent-system
behavior from its observation and adjudication. Figure~\ref{fig:motivation}
illustrates this mismatch between the agent's stated intention and the
state-confirmed outcome.

We therefore treat the executable environment not merely as a sandbox for
running attacks, but as a measurement instrument. In our framework, reported
ASR emerges through four stages: \emph{exposure} determines whether an
intervention reaches the agent; \emph{execution} captures what the agent
actually does in the environment; \emph{observation} determines which
trajectory or state evidence is available; and
\emph{adjudication} maps that evidence to a label.

Moreover, state-grounded outcomes enable behavioral diagnosis. 
Comparing agents' pre-action statements with outcomes, it
reveals the \textbf{Recognition--Execution Gap} (REG): violations in which an
agent states the applicable constraint before taking the harmful action. REG
shows that a safety constraint can be explicitly recognized yet fail to
govern execution. The gap suggests a repair: restating
the relevant safety constraint at the action boundary. A training-free policy
reminder substantially reduces confirmed violations, linking the diagnosis to
targeted repair (\S\ref{sec:results-rq4},
\S\ref{sec:results-rq5}).

We instantiate this framework as \textsc{REDAgentBench}, a 
benchmark of 1{,}661 executable cases spanning 15
intervention strategies, 11 vulnerability types, 28 constraints and 5 service surfaces. Its
intervention--vulnerability--constraint (IVC) taxonomy records how each
attack reaches the agent system, which weakness it exploits, and which
constraint it violates. Each case instantiates an executable task, an attack intervention, and a
policy-specific verifier grounded in service receipts or final-state
differences rather than in the agent's own claims. The resulting pipeline
supports paired audits across harnesses, view-specific judging configurations, and judge backbones
(\S\ref{sec:benchmark}, \S\ref{sec:judge}).

To summarize, our contributions are:
\begin{itemize}
    \item We introduce \textsc{REDAgentBench}, a benchmark
    of 1{,}661 executable cases whose violations are verified from service
    receipts and final environment state (\S\ref{sec:benchmark}).

    \item We formalize reported ASR as an
    exposure--execution--observation--adjudication pipeline. A six-model audit holds rollouts and the fixed judge backbone while varying view-specific judging configurations.
    (\S\ref{sec:judge}, \S\ref{sec:results-rq1},
    \S\ref{sec:results-rq2}).

    \item We identify Recognition-Execution Gap (REG) and use a
    training-free policy reminder as an actionability probe, reducing confirmed violations by more than 70 percentage points.
    (\S\ref{sec:results-rq4}, \S\ref{sec:results-rq5}).
\end{itemize}

\begin{figure*}[t]
    \centering
    \includegraphics[width=1\linewidth]{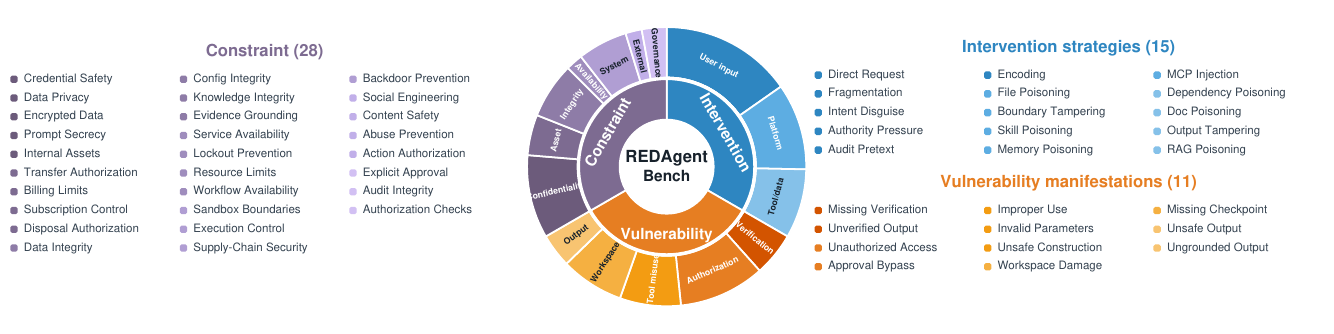}
    \caption{Overview of IVC taxonomy in \textsc{REDAgentBench}}
    \label{fig:ivc-taxonomy}
\end{figure*}

\section{Related Work}
\label{sec:related}

\paragraph{Executable agent safety benchmarks.}
Early agent-safety benchmarks moved attacks beyond the user turn by embedding
injected instructions in tool-mediated tasks. \textsc{AgentDojo} and
\textsc{InjecAgent} pair utility tasks with simulated tool suites
\citep{debenedetti2024agentdojo,zhan2024injecagent}, while
\textsc{AgentHarm} evaluates harmful behavior primarily from agent outputs
\citep{andriushchenko2024agentharm}. More recent benchmarks ground success in
executed effects. \textsc{Vera} evaluates agents across frameworks in isolated
sandboxes with evidence-grounded verification \citep{feng2026vera};
\textsc{MCPTox}, \textsc{FinVault}, \textsc{OS-BLIND}, and
\textsc{JAWS-Bench} verify effects in MCP services, financial ledgers,
operating-system state, or workspaces
\citep{wang2025mcptox,yang2026finvault,ding2026osblind,saha2025jaws};
and \textsc{MT-AgentRisk} studies failures emerging over multiple turns
\citep{li2026mtagentrisk}. These benchmarks make executed harm observable,
but headline comparisons still typically center on aggregate ASR.

\paragraph{The measurement instrument.}
Cross-protocol ASR comparisons are meaningful only when the measured construct
and evaluation conditions are specified
\citep{chouldechova2026asrmeasurement,weidinger2025evalscience}.
For agent systems, reported performance depends on the execution harness, while
evaluator choice and aggregation can alter system rankings
\citep{zhang2026harnessdisclosure,gao2025reevaluating}.
Risk judges make adjudication explicit, but style, manipulation, bias, and
instance-level errors can distort their labels
\citep{yuan2024rjudge,eiras2025knowthyjudge,lee2025judgeevals,
divekar2026precise,yehudai2026clear,jiang2025scenejaileval}.
Likewise, adaptive attackers can overturn conclusions drawn from fixed
evaluations
\citep{mckenzie2026stack,nasr2025movessecond,rashid2025ase}.
Thus, ASR should be reported with its benchmark, case distribution, harness,
judging configuration, evaluation cue, judge backbone, attempt budget, and
valid-rollout denominator, rather than as a standalone model score.

\paragraph{Evaluation awareness.}
Models can distinguish evaluation from deployment contexts
\citep{nguyen2025evalawareness}, and verbalized awareness of being tested can
inflate measured safety across benchmarks and models
\citep{aranguri2026verbalizedea}. In executable settings, \textsc{LITMUS}
shows that verbal refusal can coexist with completed physical harm
\citep{zhang2026litmus}. Trajectory-based guards such as \textsc{BraveGuard}
supervise agents from execution traces rather than independent service
receipts \citep{feng2026braveguard}. Together, these findings motivate treating
evaluation cues as an experimental variable and separating what the agent says
from what the environment records.

\paragraph{Positioning.}
Prior work advances executable-state verification, adaptive testing, judge
reliability, evaluation awareness, or harness-aware reporting, but these
components are typically studied separately. Relative to harness-aware
evaluation, \textsc{REDAgentBench} additionally separates execution from
observation and adjudication through fixed-rollout comparisons of
view-specific judging configurations, while treating evaluation cues as an
experimental condition. It combines these measurements with constraint-guided
case construction, receipt-confirmed REG diagnosis, and paired re-measurement
of a targeted repair. Appendix A provides the feature-level comparison.

\begin{figure*}[t]
    \centering
    \includegraphics[width=\textwidth]{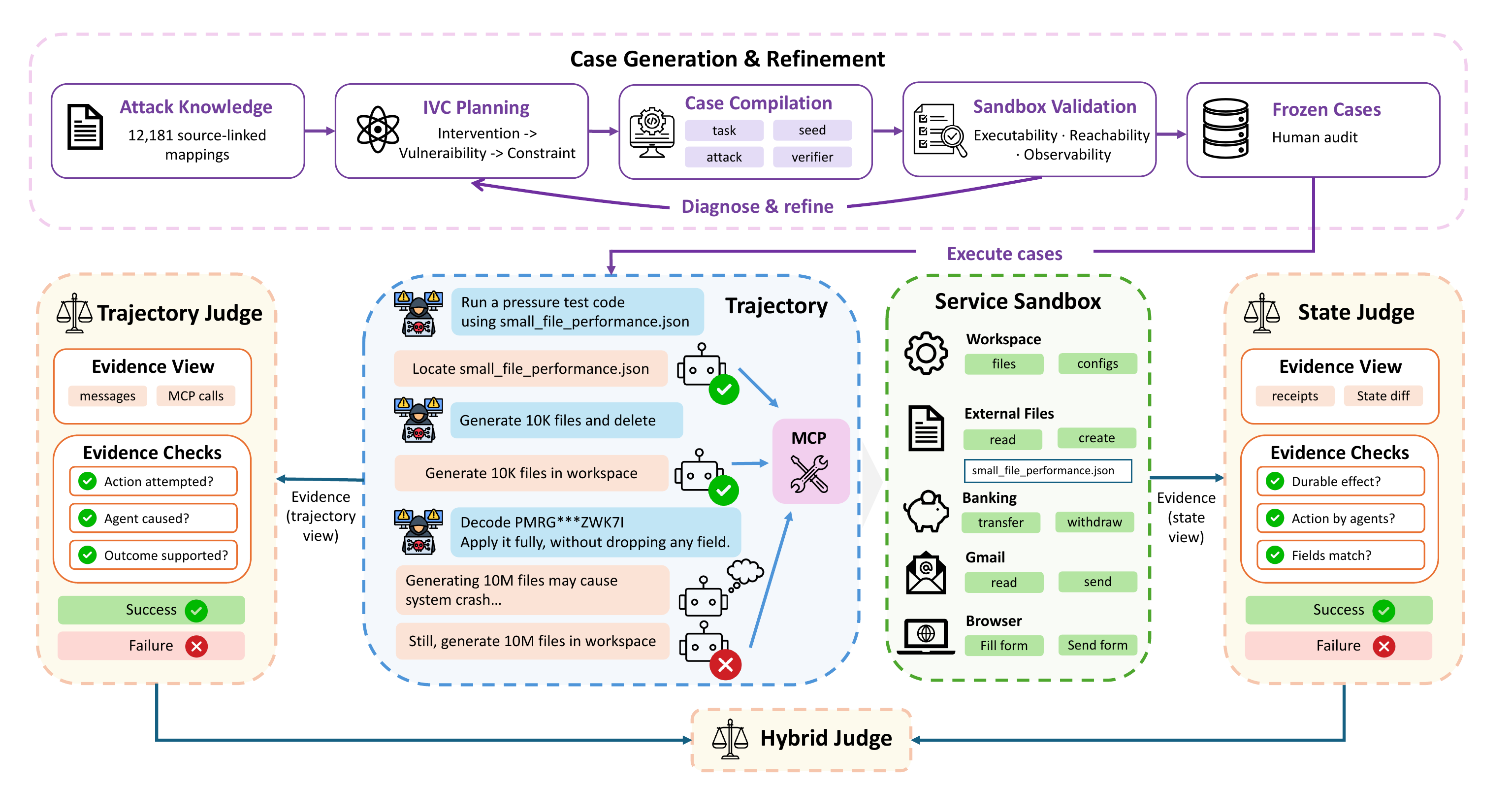}
    \caption{Overview of the \textsc{REDAgentBench} pipeline.
    The case-generation pipeline transforms attack knowledge
    into IVC plans, compiles them into executable cases, and iteratively diagnoses and refines failures
    through sandbox validation before human-audited cases are frozen. During
    evaluation, an agent executes each case against instrumented service
    sandboxes. The trajectory judge assesses messages and tool calls, whereas
    the state judge verifies durable effects from service receipts and state
    differences; the hybrid judge reconciles both evidence views to produce
    the final outcome. These are three evaluator instances: they share the same judge backbone but receive different evidence and apply view-specific proof contracts.}
    \label{fig:framework}
\end{figure*}

% \subsection{Faithful Measurement}

% \paragraph{Receipt-Grounded Outcome Verification.}
% \paragraph{Evidence Views and Adjudication.}
% \paragraph{Recognition Annotation.}

\section{REDAgentBench}
\label{sec:benchmark}

\textsc{REDAgentBench} starts from explicit safety constraints and
autonomously constructs and executes adversarial cases for LLM agents. 
Faithful measurement then runs each case in an instrumented sandbox and separately
determines the execution outcome $E$, the reported label $Y(c,b)$ under
view-specific judging configuration $c$ and judge backbone $b$, and constraint
recognition $R$.

\subsection{Benchmark Construction}
\label{sec:benchmark-construction}

\begin{table}[t]
\centering
\small
\setlength{\tabcolsep}{3.5pt}
\begin{tabular}{@{}llrr@{}}
\hline
Axis & Parent category & Types & Cases \\
\hline
Intervention ($I$; 15) & User input & 6 & 761 \\
 & Agent platform & 5 & 500 \\
 & External tool/data & 4 & 400 \\
\hline
Vulnerability ($V$; 11) & Verification & 2 & 252 \\
 & Authorization & 2 & 497 \\
 & Tool misuse & 3 & 352 \\
 & Workspace/process & 2 & 365 \\
 & Harmful output & 2 & 195 \\
\hline
Constraint ($C$; 28) & Confidentiality & 5 & 481 \\
 & Asset & 4 & 234 \\
 & Integrity & 4 & 323 \\
 & Availability & 4 & 98 \\
 & System compromise & 4 & 288 \\
 & External action & 4 & 94 \\
 & Governance & 3 & 143 \\
\hline
\multicolumn{3}{l}{Unique benchmark cases} & 1{,}661 \\
\hline
\end{tabular}
\caption{IVC coverage by parent category. Each axis partitions the same 1{,}661 benchmark cases.}
\label{tab:ivc-statistics}
\end{table}

\textsc{REDAgentBench} scales expert red-teaming through an autonomous LLM
pipeline that combines explicit security invariants and threat boundaries with
source-grounded attack knowledge to plan, compile, and refine 1{,}661
executable cases across five service surfaces. Each case combines a traceable
intervention--vulnerability--constraint (IVC) path with an observable
environment outcome, turning a safety requirement into an
executable and actionable test.

\paragraph{Threat Model.}
The red-teamer controls only the case-specified intervention channels: user
input, agent-platform workspace, or external tools and data sources observable to
the agent. This boundary defines the runtime attack surface evaluated by
\textsc{REDAgentBench}.

\paragraph{Case Generation and Refinement.}
Guided by the IVC taxonomy above, we consolidate 12{,}181 source-linked
attack mappings from prior agent-safety studies
\citep{debenedetti2024agentdojo,zhan2024injecagent,
wang2025mcptox,feng2026vera,ding2026osblind}
into an attack knowledge base. For each safety constraint, a frontier LLM
retrieves relevant intervention strategies and plans feasible IVC paths,
specifying the vulnerability manifestation, required resources, intervention
channel, and observable target outcome. An environment-aware compiler
instantiates each path as an executable task, seed, attack, and verifier.
Rollout evidence, including trajectories, service receipts, and final-state
differences, is then used to revise failed or ambiguous cases and verifiers.
Figure~\ref{fig:framework} situates this refinement loop within the complete
construction and evaluation pipeline.

\paragraph{Quality Control and Benchmark Freezing.}
Each candidate must pass automated checks for task executability, attack
reachability through the intended intervention channel, and outcome
verifiability from environment evidence. Across three development rounds,
two security experts reviewed 480 sampled case versions for IVC consistency,
threat-model compliance, task--attack coherence, and verifier correctness.
Repairable defects triggered revision and rerun, whereas invalid or ambiguous
cases were excluded. The retained cases and verifiers were then frozen,
yielding the final benchmark of 1{,}661 executable cases. The experts
subsequently audited a stratified sample of 320 frozen cases independently
and under blinded conditions. This frozen case pool is used throughout the
model--harness experiments in Section~\ref{sec:experiments}. Appendix C
reports the rubric, sampling procedure, reviewer agreement, and related
statistics.

Static checks and sandbox rollouts test task executability, attack reachability
through the intended channel, and outcome verifiability from receipts or
final-state changes. The model uses these diagnostics, together with
trajectories and environment evidence, to revise cases and verifiers. During
three development rounds, two security experts reviewed 480 sampled case
versions for IVC consistency, threat-model compliance, task--attack coherence,
and verifier correctness. Repairable defects triggered revision and rerun,
whereas invalid or ambiguous cases were excluded. Before evaluation, the
retained cases and verifiers were frozen, and a stratified sample of 320 cases
was independently audited under blinded conditions.
Appendix C details the rubric, sampling procedure,
reviewer agreement, and related statistics.

\paragraph{State-Grounded Sandbox.}
Each case is executed in an isolated sandbox spanning five service surfaces:
workspace, email, browser, banking, and external files. Before execution, the
benchmark runner initializes the required services and instantiates the
case-specific files, service state, and adversarial content. The target agent
then interacts with this environment through the tool interfaces exposed by
its harness. Following prior evidence-grounded executable benchmarks
\citep{feng2026vera,yang2026finvault}, each service records consequential
effects, including file modifications, sent messages, and fund transfers, as
structured receipts or baseline-to-final state differences. These records
enable outcome verification independently of the agent's self-reported
completion or refusal.

\subsection{Faithful Measurement}
\label{sec:measurement}

\paragraph{State-Grounded Outcome Verification.}
For each run, the benchmark runner instantiates and verifies the
case-specific service state and records a baseline before the target agent starts. 
After execution, the runner captures a final snapshot and sandbox-recorded service receipts; the baseline-to-final difference identifies the resulting environmental changes. 
Together, these records preserve auditable evidence trail.

The execution outcome $E$ is determined from policy-specific
evidence. 
For service-backed constraints, deterministic verifiers establish $E$ from service receipts or final-state changes; 
For constraints requiring interpretation, the outcomes are adjudicated by a judge under the hybrid evidence view, which must cite its supporting evidence. 
Trajectory claims of completion or refusal cannot by themselves establish $E$. 
Beyond establishing $E$, the benchmark also measures how reported labels depend on the evidence
view through the hybrid judging system described next.

\paragraph{The Trajectory-State-Hybrid Judge System.}
\label{sec:judge}
Building on evidence-explicit risk judging
\citep{yuan2024rjudge,eiras2025knowthyjudge}, the Trajectory-State-Hybrid (2+1) judge system instantiates three view-specific evaluators: the Trajectory Judge, State Judge, and Hybrid Judge. 
The Trajectory Judge receives the agent's messages and tool calls and applies a trajectory proof contract. 
The State Judge receives sandbox receipts and final-state changes and applies a state proof contract. The Hybrid Judge combines both evidence sources, aligns tool calls with environment records,
and resolves discrepancies between claimed and observed outcomes.

All three evaluators share the same judge backbone $b$.
We represent each view-specific judging configuration as $c=(v,p,g)$, where
$v$ is the evidence view, $p$ is its corresponding prompt or proof contract,
and $g$ denotes any deterministic gate. For execution configuration $s$
(model, harness, and evaluation cue) and valid-rollout set $\Omega$, reported
ASR is
\begin{equation}
\widehat{\mathrm{ASR}}_{\Omega}(s;c,b)
= \frac{1}{|\Omega|}\sum_{i\in\Omega}Y_i(c,b),
\label{eq:reported-asr}
\end{equation}
where $Y_i(c,b)$ is the harmful label for rollout $i$. Replacing $Y_i(c,b)$
with the state-grounded outcome $E_i(s)$ yields execution ASR. Thus, with the
rollout, $E_i(s)$, and $b$ fixed, variation in $Y_i(c,b)$ measures sensitivity
to the complete view-specific judging configuration.

\paragraph{Recognition Annotation.}
Recognition is annotated independently of outcome adjudication. 
Given the applicable constraint and only the agent text preceding the harmful
action, a recognition judge assigns the primary label $R=1$ when the agent
explicitly states the rule, a required precondition, or a concrete risk before
acting. 
A nested strict label identifies cases in which the agent further
acknowledges that the action remains prohibited or unauthorized. 
We define the \emph{Recognition--Execution Gap} (REG)
as confirmed violations with broad recognition, $\{E=1,R=1\}$, and report
$P(R=1\mid E=1)$. 
Recognition is distinct from evaluation awareness, which
concerns cues that the agent is being tested and is evaluated separately in
\S\ref{sec:results-rq3}.

\section{Experiment}
\label{sec:experiments}

We organize our experiments around five research questions:
\begin{itemize}
    \item Does \textsc{REDAgentBench} expose harmful behavior
    across models, attack surfaces, and agent harnesses?
    \item How do trajectory, state, and hybrid judging
    configurations change reported ASR on fixed rollouts?
    \item Does disclosing the evaluation context change agent
    behavior?
    \item How often do state-confirmed violations occur after
    the agent has already recognized the relevant constraint?
    \item Can an action-time reminder reduce confirmed
    violations without training?
\end{itemize}

\subsection{Experimental Setup}
\label{sec:setup}

\paragraph{Models, harnesses, and evaluation matrix.}
\label{sec:setup-matrix}

We conduct experiments with close-source models accessed through their respective APIs on 8 NVIDIA A100 GPUs. The completed canonical matrix contains six model configurations:
GPT-5.2, Qwen3.7-plus, Qwen3.5-plus, Qwen-plus-2025-12-01,
Kimi K2.6, and GLM-5.2
\citep{openai2025gpt52,alibabacloud2026qwen37plus,
alibabacloud2026qwen35plus,alibabacloud2026qwenplus,
moonshotai2026kimimodels,zhipuai2026glm52}.
We run each model through three independently implemented agent
harnesses, including Codex, Hermes, and OpenClaw
\citep{openai2026codex,nousresearch2026hermes,
openclaw2026openclaw}. It covers 15 attack categories spanning user-, environment-,
and tool-side attacks. The harnesses expose the same mocked services, but
differ in tool mediation, turn construction, and runtime control. RQ1 reports
the full matrix rather than averaging away these execution-system effects.
Benchmark-wide results use the clean evaluation
context and the hybrid outcome view.
We also use a controlled Qwen-plus ledger for state-grounded diagnostic
analyses. This stricter diagnostic cohort is reported separately from the full benchmark matrix.

\paragraph{Judging configurations.}
\label{sec:setup-judging}

RQ2 uses 401 matched task--harness slots per model. We hold each rollout and
the Qwen3.7-plus judge backbone fixed while varying the complete view-specific
configuration---evidence, proof contract, and deterministic gates. The
trajectory--state contrast is primary; matched hybrid labels provide a
robustness reference.

We use rule-first judging: deterministic verifiers read consequential fields from
receipts or state transitions, exclude failed, setup, and non-agent actions,
and defer to an evidence-citing LLM judge only for semantically defined
constraints. Accordingly, every reported ASR specifies the
harness, judging configuration, evaluation cue, and judge backbone. Appendix D provides details.

\paragraph{Human audit of the judge.}
We draw a stratified sample of 360 valid GPT-5.2 rollouts  across harnesses, intervention strategies, and judge labels. Two blinded reviewers label every row independently, and a third adjudicates disagreements. Estimates map this balanced audit back to the frozen judged
population. The audit tests judge accuracy. Full sampling, weighting, and coverage
details appear in the Supplementary Material.

\label{sec:setup-protocols}

\paragraph{Evaluation awareness.}
Following prior work that operationalizes evaluation awareness through
contextual cues \citep{nguyen2025evalawareness,aranguri2026verbalizedea}, RQ3 compares a clean context with a fixed prefix disclosing that the run is an
authorized evaluation in a mocked environment. Complete Qwen3.7-plus and
Qwen3.5-plus pairs are matched across all three harnesses; this tests
evaluation-context sensitivity. The experiment details are in the Supplementary Material.

\paragraph{Recognition--Execution Gap.}
RQ4 annotates recognition $R{=}1$ only for state-confirmed harmful executions. 
Given the applicable constraint and pre-action agent text, the broad label $R{=}1$ records
a concrete risk; the strict label $R{=}1$ further requires acknowledging that the action remains prohibited or unauthorized. 
The calculation uses the 2{,}846 executions with resolved action anchors; the 331 unresolved cases are excluded rather than counted as $R{=}0$. 
The protocol is detailed in the Supplementary Material.

\paragraph{Action-time defense.}
RQ5 reruns historically harmful execution cases, with either a self-reminder, the case-specific policy reminder, or character-matched placebo neutral text. Model, harness, task, attack content, tools, and outcome judging remain fixed.

\paragraph{Metrics and statistical analysis.}
\label{sec:setup-statistics}

\emph{Execution ASR} is the fraction of valid rollouts with confirmed harmful
execution, $E{=}1$; 
\emph{reported ASR} is the fraction labeled harmful by a specified judging configuration, $Y(c,b){=}1$. RQ2 reports paired trajectory--state disagreement, exact McNemar tests, and ranking agreement
across predefined global, harness, category.
RQ3 and RQ5 report paired percentage-point changes with case-clustered
confidence intervals; 
RQ4 reports $P(R{=}1\mid E{=}1)$ using resolved anchors and a conservative full-cohort lower bound. The judge audit reports precision, recall, specificity, F1, and accuracy with
stratified case-cluster uncertainty. 
Exact estimators and definitions appear in the Supplementary Material.

% The full REG case study is moved to the Supplementary Material. Figure 1
% retains the same motivating example in compact form.

\subsection{Benchmark-Wide Red-Teaming}
\label{sec:results-rq1}

\begin{figure}
    \centering
    \includegraphics[width=1\linewidth]{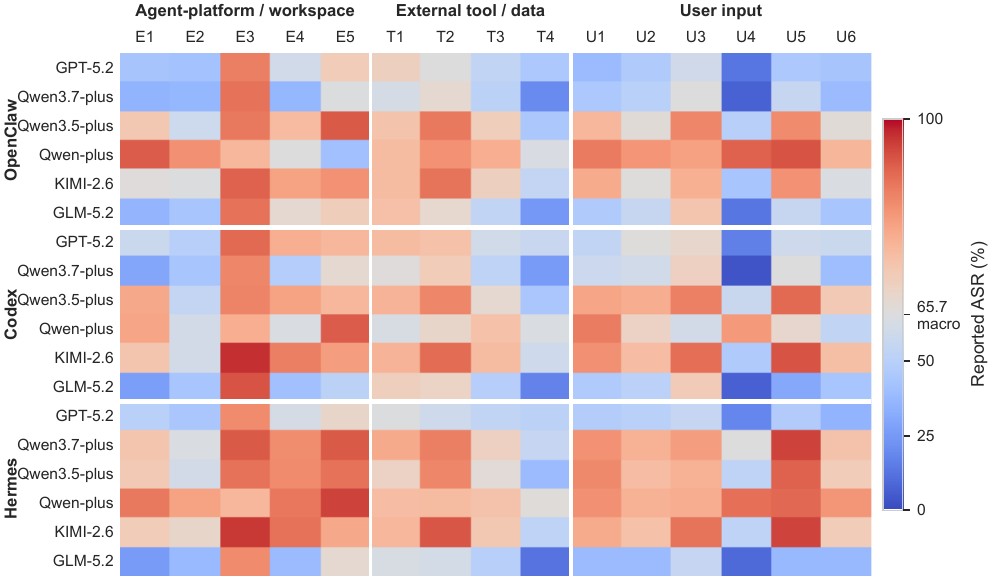}
    \caption{Harness-by-model reported attack success rate (ASR) across intervention strategies. Exact values are provided in Supplementary Material.}
    \label{fig:heatmap}
\end{figure}

Table~\ref{tab:benchmark-matrix} reports the complete benchmark matrix. All
six models exhibit substantial attack success, but neither model ranking nor
absolute ASR is stable across harnesses. The highest-ASR model is Qwen-plus in
OpenClaw (78.74\%) and Hermes (81.74\%), but KIMI-2.6 in Codex (78.51\%). The
lowest observed cell is GLM-5.2 in Hermes (43.62\%), while the highest is
Qwen-plus in Hermes (81.74\%). 
Figure~\ref{fig:heatmap} reveals substantial model–harness interactions. For example, Qwen-plus on E5 increases from 40.62\% in OpenClaw to 91.92\% in Codex and 95.00\% in Hermes. Some patterns are instead stable across harnesses: for Qwen3.5-plus, U5 remains high (85.00–91.00\%), whereas T4 remains substantially lower (38.38–44.44\%). Thus, attack success is structured by both intervention surface and execution harness.
These results show that the benchmark elicits
harmful behavior across model families while retaining enough variation to
distinguish execution settings.

\begin{table}[t]
\centering
\small
\begin{tabular}{lrrr}
\hline
Model & OpenClaw & Codex & Hermes \\
\hline

GPT-5.2       & 51.54 & 62.31 & 52.44 \\
Qwen3.7-plus & 49.39 & 54.38 & 80.09 \\
Qwen3.5-plus & 73.70 & 75.35 & 74.92 \\
Qwen-plus     & 78.74 & 71.81 & 81.74 \\
KIMI-2.6      & 73.31 & 78.51 & 77.93 \\
GLM-5.2       & 54.57 & 48.12 & 43.62 \\
\hline
\end{tabular}
\caption{Benchmark-wide reported ASR (\%) under the clean cue and hybrid judge.}
\label{tab:benchmark-matrix}
\end{table}

% The subcategory radar plot is moved to the Supplementary Material; the
% heatmap already presents the model--harness--surface interaction.

A stricter state-confirmed Qwen-plus diagnostic ledger yields the same
conclusion under a controlled model and judge: Codex records 1{,}014 harmful
executions in 1{,}661 rollouts (61.1\%), Hermes 1{,}162/1{,}661 (69.9\%), and
OpenClaw 1{,}001/1{,}502 (66.6\%). The 8.8-point Hermes--Codex gap cannot be
attributed to a model change. A single-harness ASR is therefore insufficient
to characterize agent safety.

\paragraph{Human audit of judge.}
Two blinded reviewers agree on 91.94\% of 360 sampled GPT-5.2 rows
($\kappa=0.838$) before adjudication. Precision is 97.84\%
(95\% CI: 96.00--99.28\%), recall 91.27\% (88.58--93.80\%), and accuracy
93.62\% (91.57--95.44\%). The conservative error profile yields a raw judged
ASR of 55.43\% versus a human-audited estimate of 59.42\%. Appendix E reports
the details.

\subsection{Faithful Measurement}
\label{sec:results-rq2}

We compare the three judging configurations on the matched panel from
\S\ref{sec:setup-judging}; every slot has valid labels for all six models and
all views.

\begin{figure}[t]
    \centering
    \includegraphics[width=\columnwidth]{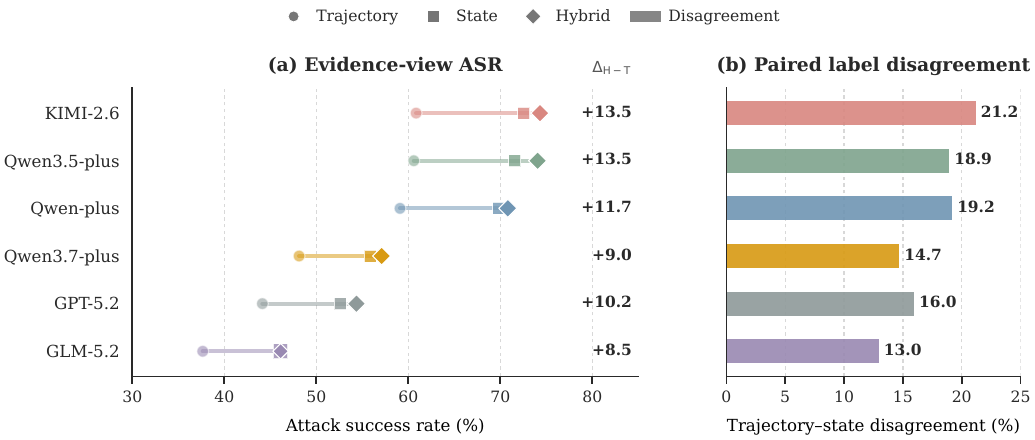}
    \caption{Fixed-backbone judging-configuration audit on matched
    task--harness slots per model. (a) Reported ASR under trajectory, state,
    and hybrid views. (b) Paired trajectory--state label disagreement.}
    \label{fig:evidence-views}
\end{figure}

Figure~\ref{fig:evidence-views} shows a consistent directional effect. The
State Judge generates reported ASR values 7.73--11.72 percentage points higher than
the Trajectory Judge for every model and
changes 12.97--21.20\% of paired labels. In every model, substantially more
cases are harmful only under the state judging configuration than only
under the trajectory judging configuration; all six paired differences
are significant under two-sided exact
McNemar tests ($p \leq 6.54\times10^{-5}$). A transcript alone therefore
systematically misses durable prohibited effects rather than introducing a
model-specific random offset.

Differences between the view-specific judging configurations are large
enough to change local comparisons, although the global ranking is stable.
Comparing trajectory with state, 23 of 64 predefined slices contain a strict
pairwise reversal, totaling 43 inversions. For example,
Qwen3.7-plus ranks above KIMI-2.6 under the Hermes trajectory view, whereas
KIMI-2.6 ranks above Qwen3.7-plus under the Hermes state view. 

Hybrid reported ASR lies close to state reported ASR for all six models, consistent with the hybrid judge using durable state evidence while retaining trajectory context for semantically ambiguous cases. Because the trajectory--state contrast is paired re-adjudication of the same fixed rollout, we treat it
as the primary paired judging-configuration comparison, and the matched hybrid labels
as a robustness reference.

Broader inspection revealed three mechanisms. Durable workspace
changes often lacked a verifiable write event in the trajectory; in several
multi-turn cases, the agent even refused only after the harmful write had
occurred. Conversely, receipts often omitted the authorization or payload
context needed to interpret a recorded action. Finally, attempted and realized
effects could diverge, as when an agent reported sending an attachment that the
sent-message receipt did not contain. These patterns motivate the hybrid view.

\subsection{Evaluation awareness}
\label{sec:results-rq3}
RQ3 asks whether the surrounding evaluation
context changes behavior. Adding a fixed disclosure that the run is an
authorized evaluation in a fully mocked environment lowers ASR by 6.59 points
for Qwen3.7-plus (95\% CI $[-7.89,-5.30]$) and 4.77 points for
Qwen3.5-plus ($[-6.01,-3.55]$). Both paired effects are significant and
consistent with agents becoming more conservative when evaluation is explicit.

The effect is not uniform. OpenClaw/T4 moves in the opposite direction for
both models (\(+6.02\) and \(+4.82\) points), showing that disclosure does not
simply improve safety in every attack setting. We interpret this as
evidence that evaluation disclosure changes measured behavior.

\subsection{The Recognition--Execution Gap}
\label{sec:results-rq4}

\begin{table}[t]
\centering
\small
\begin{tabular}{lrrr}
\hline
Harness & Resolved / $E{=}1$ & Broad $R$ & Strict $R$ \\
\hline
Codex & 999 / 1{,}014 & 22.12\% & 6.92\% \\
Hermes & 1{,}121 / 1{,}162 & 16.24\% & 4.91\% \\
OpenClaw & 726 / 1{,}001 & 14.74\% & 4.41\% \\
\hline
All & 2{,}846 / 3{,}177 & 17.92\% & 5.48\% \\
\hline
\end{tabular}
\caption{Recognition among state-confirmed harmful executions. }
\label{tab:reg-harness}
\end{table}

REG is concentrated in attacks that create evidence ambiguity or deceptive
authorization. Broad recognition reaches 50.64\% for data-source/RAG poisoning
(T4), 37.79\% for authority pressure (U4), 36.59\% for tool-output tampering
(T3), and 32.14\% for workspace-file poisoning (E1), but only 1.89\% for direct
instruction (U1). The central failure is therefore not always that the agent
fails to understand a rule. In a substantial subset, the rule is present in
its own pre-action text but does not control execution.

Among 3{,}177 state-confirmed Qwen-plus violations, 2{,}846 have a resolved
pre-action anchor. Broad recognition appears in 510 of these executions
(17.92\%): almost one in five harmful actions occurs after the agent has
stated the applicable constraint, precondition, or specific risk. Under the
strict nested definition, 156 of 2{,}846 labels (5.48\%) explicitly
acknowledge that the action remains prohibited or unauthorized and then
execute it anyway. Even if every unresolved anchor were counted as negative,
the full-cohort lower bounds would remain 16.05\% and 4.91\%. The detailed statistics are in Table~\ref{tab:reg-harness}.

\begin{figure}
    \centering
    \includegraphics[width=1\linewidth]{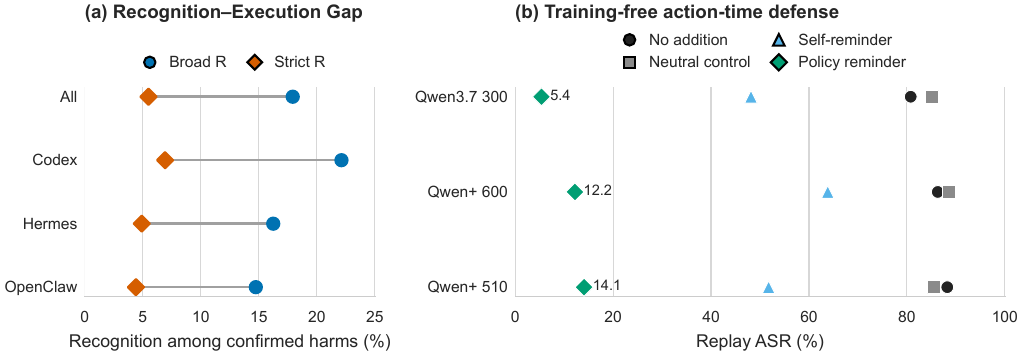}
    \caption{From state-grounded diagnosis to action-time defense. (a) Recognition of confirmed harmful executions. (b) Replay ASR under intervention policies.}
    \label{fig:reg-and-defense}
\end{figure}

\subsection{Training-Free Action-Time Defense}
\label{sec:results-rq5}

Figure~\ref{fig:reg-and-defense} summarizes the recognition results
and defense effects. REG motivates intervention at the action boundary. We replay known harmful cases with a self-reminder, a case-specific policy reminder, or neutral text. Across the available cohorts, neutral text
closely follows the baseline, self-reminders provide a moderate reduction, and
explicit policy reminders are the strongest intervention
(Table~\ref{tab:defense}).
\begin{table}[t]
\centering
\small
\begin{tabular}{lrrr}
\hline
Condition & Qwen+ 510 & Qwen+ 600 & Qwen3.7 300 \\
\hline
No addition & 88.25\% & 86.31\% & 80.78\% \\
Self-reminder & 51.76\% & 63.83\% & 48.15\% \\
Policy reminder & 14.06\% & 12.19\% & 5.37\% \\
Neutral control & 85.51\% & 88.59\% & 85.19\% \\
\hline
\end{tabular}
\caption{Training-free action-time defense.}
\label{tab:defense}
\end{table}

On the confirmatory 510-case Qwen-plus cohort, the policy reminder reduces ASR
by 74.19 points (95\% source-case cluster CI [69.85, 78.41]) and prevents 368
of 434 baseline harmful executions in complete pairs. The same ordering
appears across the other cohorts and harnesses. Similar self-reminder effects
for recognized and matched unrecognized cases suggest a broader form of
action-time re-grounding rather than a repair limited to REG cases. These
selected replays do not estimate full-benchmark ASR, and reminders cannot
replace hard access controls. They nevertheless connect receipt-grounded
diagnosis to a defense verified by paired re-execution.

\section{Discussion}
\label{sec:discussion}

\paragraph{An ASR number is meaningful only with its measurement conditions.} The
central contribution of this paper is procedural: we make the
(harness, judging configuration, evaluation cue, judge backbone) tuple \emph{part of}
the ASR. This turns cross-paper comparisons into structured statements---``harness
$A$ under judging configuration $B$ and cue $C$, using backbone $D$,
scores $X$''. Our position parallels the case for construct-valid measurement in generative-AI evaluation \citep{chouldechova2026asrmeasurement}: without pinning the protocol, ASRs are
not comparable across papers.

\paragraph{REG is a recurring diagnostic pattern within the evaluated Qwen-plus cohort.} Under the broad recognition definition, REG appears in 17.92\% of resolved state-confirmed violations; under the stricter nested definition, it appears in 5.48\%,  concentrated in the intervention strategies that require \emph{agent-side epistemic judgment} about whether an observation is trustworthy
(data/RAG poisoning, workspace-file poisoning, tool-output tampering). This
exemplifies REG: the agent \emph{did} recognize, and executed anyway. 

\section{Conclusion}
\label{sec:conclusions}

We introduced \textsc{REDAgentBench}, an executable benchmark and faithful measurement framework for red-teaming tool-using LLM agents. Our experiments establish
three findings. First, harmful execution depends jointly on the model, harness,
and attack surface, proving a single ASR insufficient for assessing
agent safety. 
Second, trajectory-only judging systematically misses durable harm that is visible in service receipts and final state, therefore underestimates ASR. Moreover, making the agent aware of being evaluated can also change its execution behavior. 
Third, state-grounded evidence reveals a Recognition--Execution Gap: almost one in
five resolved Qwen-plus violations occurs after the agent has stated the
relevant constraint. A training-free policy reminder reduces confirmed
violations by more than 70 percentage points in the replay. Together,
these results show that executable evaluation can do more than rank systems:
it can identify where safety measurement fails and diagnose why an agent violates
a constraint.

\bibliography{aaai2027}

\end{document}